\documentclass[final]{cvpr}

\usepackage{times}
\usepackage{epsfig}
\usepackage{graphicx}
\usepackage{amsmath}
\usepackage{amssymb}

\usepackage{url}            
\usepackage{booktabs}       
\usepackage{amsfonts}       
\usepackage{nicefrac}       
\usepackage{microtype}      
\usepackage{multirow}
\usepackage{makecell}
\usepackage{subfigure}
\usepackage{array}
\usepackage{wrapfig}
\usepackage{xcolor}

\usepackage[pagebackref=true,breaklinks=true,colorlinks,bookmarks=false]{hyperref}

\def\cvprPaperID{4547} 
\def\confYear{CVPR 2021}

\begin{document}

\title{Dynamic Slimmable Network}

\author{Changlin Li$^1$ \quad Guangrun Wang$^2$ \quad Bing Wang$^3$ \quad Xiaodan Liang$^4$ \quad Zhihui Li$^5$ \quad Xiaojun Chang$^1$  \\
\small$^1$ GORSE Lab, Dept. of DSAI, Monash University \quad
\small$^2$ Univeristy of Oxford \quad
\small$^3$ Alibaba Group \\
\small$^4$ Sun Yat-Sen University \quad
\small$^5$ Shandong Artificial Intelligence, Qilu University of Technology \\
{\tt\small changlin.li@monash.edu, wanggrun@gmail.com, fengquan.wb@alibaba-inc.com,}\\{\tt\small xdliang328@gmail.com, zhihuilics@gmail.com, xiaojun.chang@monash.edu}
}

\maketitle

\begin{abstract}
    Current dynamic networks and dynamic pruning methods have shown their promising capability in reducing theoretical computation complexity. However, dynamic sparse patterns on convolutional filters fail to achieve actual acceleration in real-world implementation, due to the extra burden of indexing, weight-copying, or zero-masking. Here, we explore a dynamic network slimming regime, named Dynamic Slimmable Network (DS-Net), which aims to achieve good hardware-efficiency via dynamically adjusting filter numbers of networks at test time with respect to different inputs, while keeping filters stored statically and contiguously in hardware to prevent the extra burden. Our DS-Net is empowered with the ability of dynamic inference by the proposed double-headed dynamic gate that comprises an attention head and a slimming head to predictively adjust network width with negligible extra computation cost. To ensure generality of each candidate architecture and the fairness of gate, we propose a disentangled two-stage training scheme inspired by one-shot NAS. In the first stage, a novel training technique for weight-sharing networks named In-place Ensemble Bootstrapping is proposed to improve the supernet training efficacy. In the second stage, Sandwich Gate Sparsification is proposed to assist the gate training by identifying easy and hard samples in an online way. Extensive experiments demonstrate our DS-Net consistently outperforms its static counterparts as well as state-of-the-art static and dynamic model compression methods by a large margin (up to 5.9\%). Typically, DS-Net achieves 2-4$\times$ computation reduction and 1.62$\times$ real-world acceleration over ResNet-50 and MobileNet with minimal accuracy drops on ImageNet.\footnote{Code release: \url{https://github.com/changlin31/DS-Net}}
\end{abstract}

\vspace{-15pt}
\section{Introduction}
\vspace{-5pt}
\begin{figure}[t]
    \centering
    \includegraphics[width=0.8\linewidth]{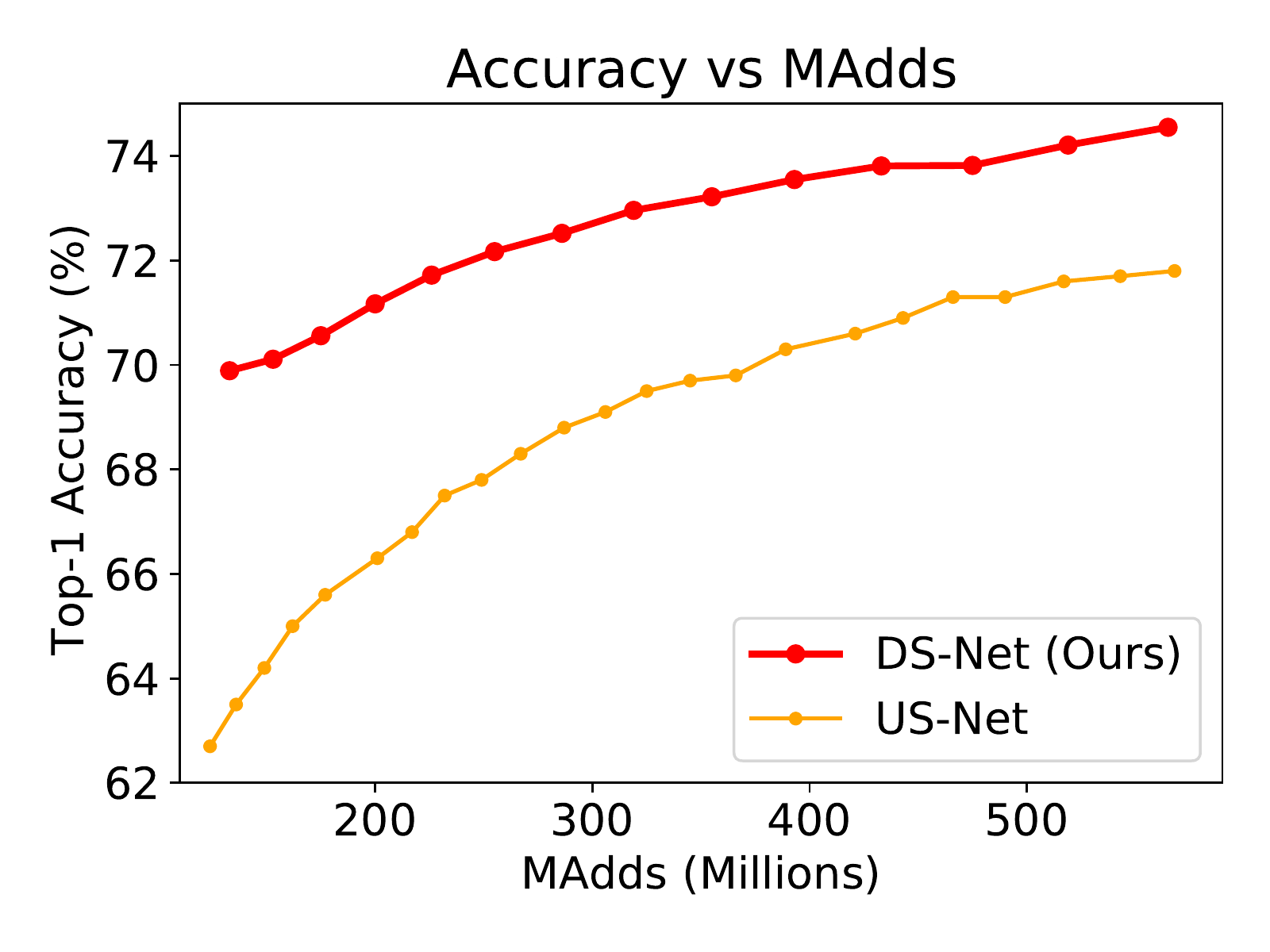}
    \vspace{-10pt}
    \caption{\small Universally accuracy-complexity comparison of our DS-Net and Universally Slimmable Network (US-Net) \cite{Yu2019UniversallySN} (based on MobileNetV1 \cite{howard2017mobilenets}).} 
    \label{fig:universal}
\vspace{-5pt}
\end{figure}
\begin{figure*}
\vspace{-10pt}
\centering
\subfigure{\includegraphics[width=0.3\linewidth]{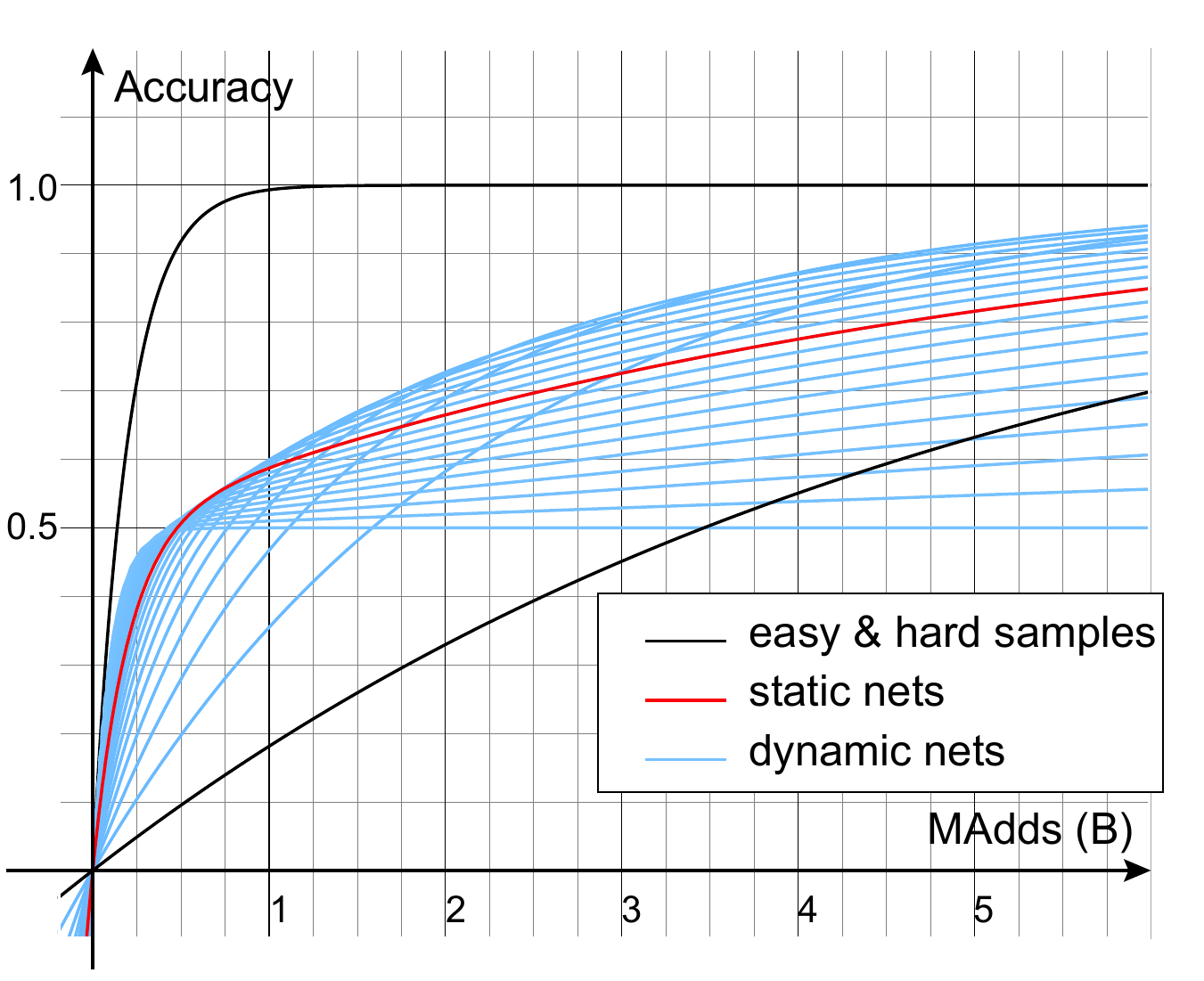}}~~~~~~~~~~~~~~~%
\subfigure{\includegraphics[width=0.5\linewidth]{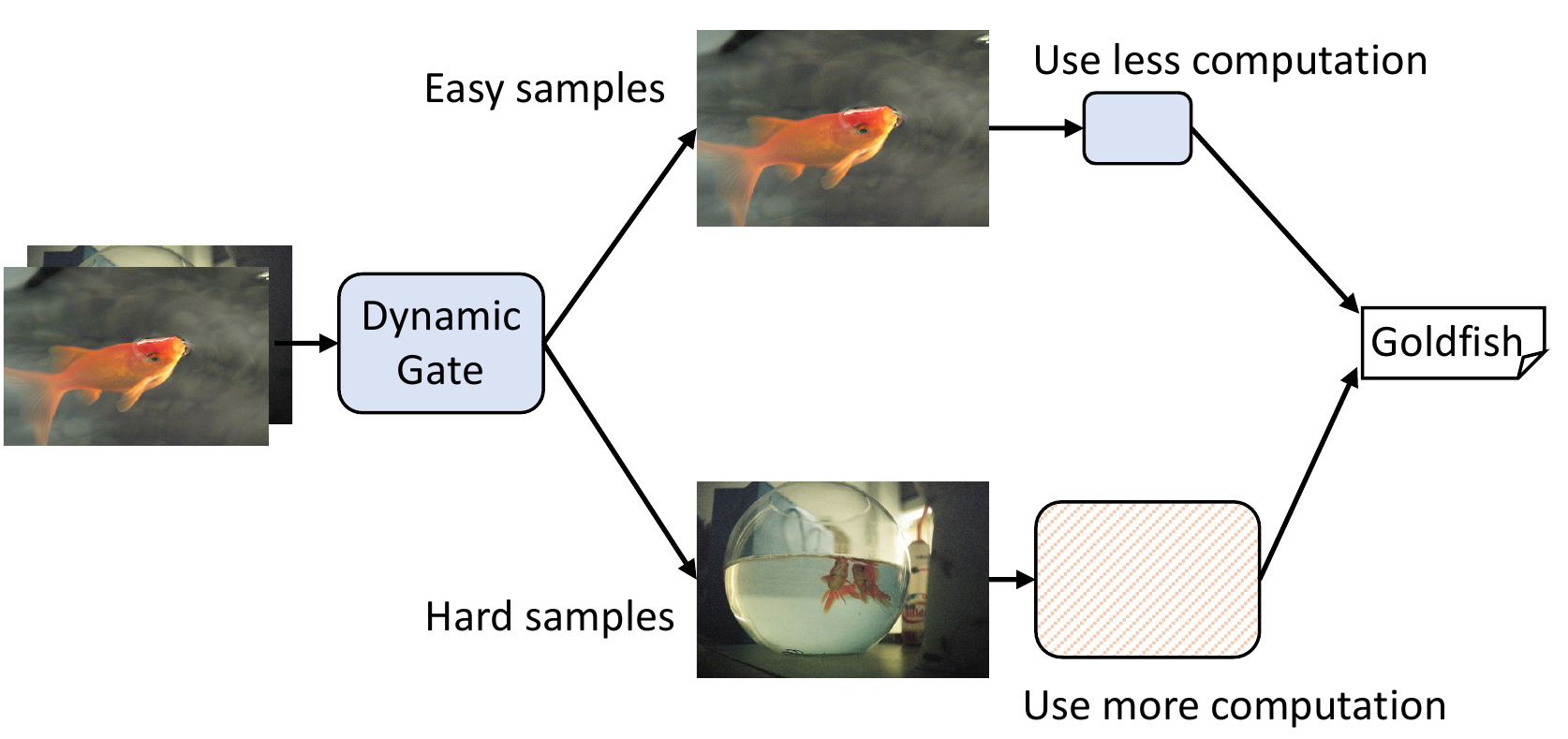}}%
\caption{
The motivation for designing dynamic networks to achieve efficient inference. 
\textbf{Left:} A simulation diagram of accuracy-complexity comparing a series of static networks (searched by NAS) with 20 dynamic inference schemes of different resource allocate proportion for easy and hard samples on a hypothetical classification dataset with evenly distributed easy and hard samples.
\textbf{Right:} Illustration of dynamic networks on efficient inference. Input images are routed to use different architectures regarding their classification difficulty.}\label{fig:motivation}\vspace{-10pt}
\end{figure*}
\begin{table}
 \caption{Latency comparison of ResNet-50 with 25\% channels (on GeForce RTX 2080 Ti).
 Both \textit{masking} and \textit{indexing} lead to inefficient computation waste, while \textit{slicing} achieves comparable acceleration with \textit{ideal} (the individual ResNet-50 0.25$\times$).
 }
 \label{tab:ds_vs_dp}
 \centering
 \footnotesize
 \begin{tabular}{llllll}
  \toprule
  method  & full & masking & indexing & slicing (ours) & ideal \\
  \midrule
  latency & 12.2 ms & 12.4ms & 16.6 ms & 7.9 ms  & 7.2 ms\\
  \bottomrule
  \vspace{-25pt}
 \end{tabular}
\end{table}
As deep neural networks are becoming deeper and wider to achieve higher performance, there is an urgent need to explore efficient models for common mobile platforms, such as self-driving cars, smartphones, drones and robots. In recent years, many different approaches have been proposed to improve the inference efficiency of neural networks, including network pruning~\cite{Li2016PruningFF, liu2017NetworkSlimming, SoftFilterPruning, He2017ChannelPF, Liu2019MetaPruningML, luo2017thinet}, weight quantization~\cite{jacob2018quantization}, knowledge distillation~\cite{Ba2013DoDN, Romero2014FitNetsHF, Hinton2015DistillingTK}, manually and automatically designing of efficient networks~\cite{Tan2019EfficientNetRM,Sandler2018MobileNetV2IR,ZhangLPCZGS21,RenXCHLCW2020,bender2018understanding, ZhangLPCGS20, ChengZHDCLDG20, Guo2019SinglePO, li2019blockwisely, ZhangLPCS20} and dynamic inference~\cite{Bolukbasi2017AdaptiveNN, Huang2018MultiScaleDN, wang2018skipnet, veit2018AIG, Li2019ImprovedTF,hua2019channel, gao2018dynamic}. 

Among the above approaches, dynamic inference methods, including networks with dynamic depth \cite{Bolukbasi2017AdaptiveNN, Huang2018MultiScaleDN, wang2018skipnet, veit2018AIG,Li2019ImprovedTF} and dynamic width \cite{gao2018dynamic,hua2019channel,Chen2019YouLT} have attracted increasing attention because of their promising capability of reducing computational redundancy by automatically adjusting their architecture for different inputs. As illustrated in Fig. \ref{fig:motivation}, the dynamic network learns to configure different architecture routing adaptively for each input, instead of optimizing the architecture among the whole dataset like Neural Architecture Search (NAS) or Pruning. A performance-complexity trade-off simulated with exponential functions is also shown in Fig. \ref{fig:motivation}, the optimal solution of dynamic networks is superior to the static NAS or pruning solution. Ideally, dynamic network routing can significantly improve model performance under certain complexity constraints.

However, networks with dynamic width, \textit{i.e.}, dynamic pruning methods~\cite{gao2018dynamic,hua2019channel,Chen2019YouLT}, unlike its orthogonal counterparts with dynamic depth, have never achieved actual acceleration in a real-world implementation. As natural extensions of network pruning, dynamic pruning methods predictively prune the convolution filters with regard to different input at runtime. The varying sparse patterns are incompatible with computation on hardware. Actually, many of them are implemented as zero masking or inefficient path indexing, resulting in a massive gap between the theoretical analysis and the practical acceleration. As shown in Tab. \ref{tab:ds_vs_dp}, both masking and indexing lead to inefficient computation waste.

To address the aforementioned issues in dynamic networks, we propose \textbf{D}ynamic \textbf{S}limmable \textbf{N}etwork (DS-Net), which achieves good hardware-efficiency via dynamically adjusting filter numbers of networks at test time with respect to different inputs. To avoid the extra burden on hardware caused by dynamic sparsity, we adopt a scheme named \textbf{dynamic slicing} to keep filters static and contiguous when adjusting the network width. Specifically, we propose a \textbf{double-headed dynamic gate} with an attention head and a slimming head upon slimmable networks to predictively adjust the network width with negligible extra computation cost. The training of dynamic networks is a highly entangled bilevel optimization problem. To ensure generality of each candidate's architecture and the fairness of gate, a disentangled \textbf{two-stage training scheme} inspired by one-shot NAS is proposed to optimize the supernet and the gates separately. In the first stage, the slimmable supernet is optimized with a novel training method for weight-sharing networks, named \textbf{In-place Ensemble Bootstrapping (IEB)}. IEB trains the smaller sub-networks in the online network to fit the output logits of an ensemble of larger sub-networks in the momentum target network. Learning from the ensemble of different sub-networks will reduce the conflict among sub-networks and increase their generality. Using the exponential moving average of the online network as the momentum target network can provide a stable and accurate historical representation, and bootstrap the online network and the target network itself to achieve higher overall performance. In the second stage, to prevent dynamic gates from collapsing into static ones in the multiobjective optimization problem, a technique named \textbf{Sandwich Gate Sparsification (SGS)} is proposed to assist the gate training. During training, SGS identifies easy and hard samples online and further generates the ground truth label for the dynamic gates. 

Overall, our contributions are three-fold as follows:\begin{itemize}
\vspace{-10pt}
\item{} We propose a new dynamic network routing regime, achieving good hardware-efficiency by predictively adjusting filter numbers of networks at test time with respect to different inputs. Unlike dynamic pruning methods, we dynamically slice the network parameters while keeping them stored statically and contiguously in hardware to prevent the extra burden of masking, indexing, and weight-copying. The dynamic routing is achieved by our proposed double-headed dynamic gate with negligible extra computation cost.
\vspace{-8pt}
\item{} We propose a two-stage training scheme with IEB and SGS techniques for DS-Net. Proved experimentally, IEB stabilizes the training of slimmable networks and boosts its accuracy by 1.8\% and 0.6\% in the slimmest and widest sub-networks respectively. Moreover, we empirically show that the SGS technique can effectively sparsify the dynamic gate and improves the final performance of DS-Net by 2\%.
\vspace{-8pt}
\item{}Extensive experiments demonstrate our DS-Net outperforms its static counterparts \cite{yu2019autoslim,Yu2019UniversallySN} as well as state-of-the-art static and dynamic model compression methods by a large margin (up to 5.9\%, Fig. \ref{fig:universal}). Typically, DS-Net achieves 2-4$\times$ computation reduction and 1.62$\times$ real-world acceleration over ResNet-50 and MobileNet with minimal accuracy drops on ImageNet. Gate visualization proves the high dynamic diversity of DS-Net.
\end{itemize}
\vspace{-10pt}
\begin{figure*}[ht]
\vspace{-10pt}
  \centering
   \includegraphics[width=0.9\linewidth]{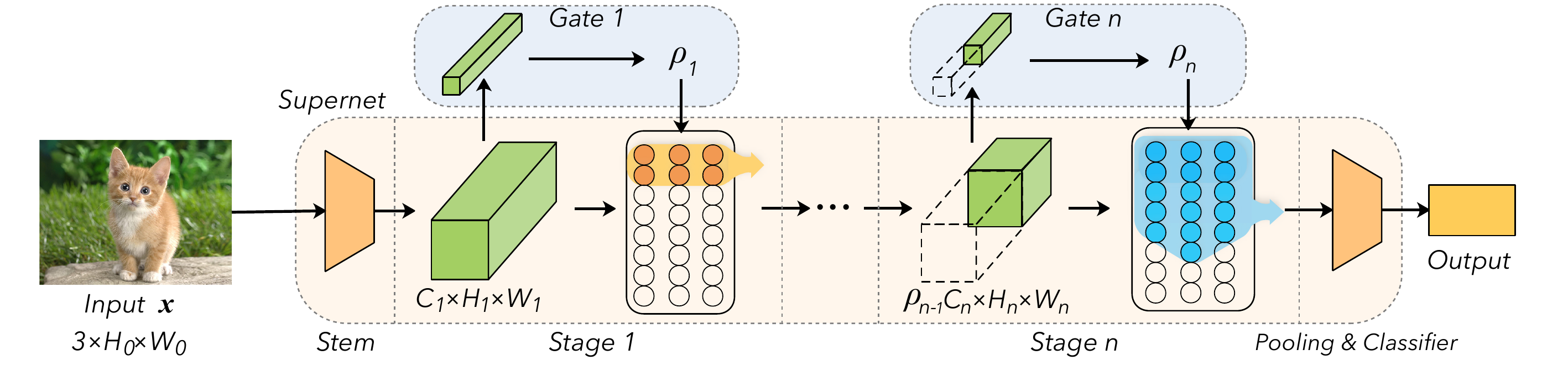}
  \caption{Architecture of DS-Net. The width of each supernet stage is adjusted adaptively by the slimming ratio $\rho$ predicted by the gate.}\label{fig:network}
  \vspace{-10pt}
\end{figure*}
\section{Related works}
\vspace{-5pt}
\noindent\textbf{Anytime neural networks} \cite{larsson2016fractalnet, Huang2018MultiScaleDN, hu2019learning, Li2019ImprovedTF, Lee2018AnytimeNP, Yu2019SlimmableNN, Yu2019UniversallySN,hou2020dynabert} are single networks that can execute with their sub-networks under different budget constraints, thus can deploy instantly and adaptively in different application scenarios. Anytime neural networks have been studied in two orthogonal directions: networks with variable depth and variable width. \textbf{Networks with variable depth} \cite{larsson2016fractalnet, Huang2018MultiScaleDN, hu2019learning, Li2019ImprovedTF} are first studied widely, benefiting from the naturally nested structure in depth dimension and residual connections in ResNet \cite{he2016deep} and DenseNet \cite{huang2017densely}. \textbf{Network with variable width} was first studied in \cite{Lee2018AnytimeNP}. Recently, slimmable networks \cite{Yu2019SlimmableNN, Yu2019UniversallySN} using \emph{switchable batch normalization} and \emph{in-place distillation} achieve higher performance than their stand-alone counterparts in any width. Some recent works \cite{cai2019once,Yu2020BigNASSU,hou2020dynabert} also explore anytime neural networks in multiple dimensions, \textit{e.g.} depth, width, kernel size, \textit{etc.}

\noindent\textbf{Dynamic neural networks} \cite{veit2018AIG,wang2018skipnet,li2020DynamicRouting,Yang2019CondConvCP} change their architectures based on the input data. Dynamic networks for efficient inference aim to reduce average inference cost by using different sub-networks adaptively for inputs with diverse difficulty levels. \textbf{Networks with dynamic depth} \cite{Bolukbasi2017AdaptiveNN, Huang2018MultiScaleDN, wang2018skipnet, veit2018AIG, Li2019ImprovedTF} achieve efficient inference in two ways, early exiting when shallower sub-networks have high classification confidence \cite{Bolukbasi2017AdaptiveNN, Huang2018MultiScaleDN, Li2019ImprovedTF}, or skipping residual blocks adaptively \cite{wang2018skipnet, veit2018AIG}. Recently, \textbf{dynamic pruning} methods~\cite{hua2019channel,gao2018dynamic,Chen2019YouLT} using a variable subset of convolution filters have been studied. Channel Gating Neural Network \cite{hua2019channel} and FBS \cite{gao2018dynamic} identify and skip the unimportant input channels at run-time. In GaterNet \cite{Chen2019YouLT}, a separate gater network is used to predictively select the filters of the main network. Please refer to \cite{han2021dynamic} for a more comprehensive review of dynamic neural networks.

\noindent\textbf{Weight sharing NAS}~\cite{brock2017smash, akimoto2019adaptive, bender2018understanding, Liu2018DARTSDA, Cai2018ProxylessNASDN, Wu2018FBNetHE, Guo2019SinglePO, li2019blockwisely,cai2019once,li2021bossnas}, aiming at designing neural network architectures automatically and efficiently, has been developing rapidly in recent two years. They integrate the whole search space of NAS into a weight sharing supernet and optimize network architecture by pursuing the best-performing sub-networks. These methods can be roughly divided into two categories: \textbf{jointly optimized methods} \cite{Liu2018DARTSDA, Cai2018ProxylessNASDN, Wu2018FBNetHE}, in which the weight of the supernet is jointly trained with the architecture routing agent (typically a simple learnable factor for each candidate route); and \textbf{one-shot methods} \cite{brock2017smash, akimoto2019adaptive, bender2018understanding, Guo2019SinglePO, li2019blockwisely,cai2019once,li2021bossnas}, in which the training of the supernet parameters and architecture routing agent are disentangled. After fair and sufficient training, the agent is optimized with the weights of supernet frozen.
\vspace{-8pt}


\section{Dynamic Slimmable Network}
\label{methods}
\vspace{-5pt}
Our dynamic slimmable network achieves dynamic routing for different samples by learning a slimmable supernet and a dynamic gating mechanism. 
As illustrated in Fig. \ref{fig:network}, 
the supernet in DS-Net refers to the whole module undertaking the main task. In contrast, the dynamic gates are a series of predictive modules that route the input to use sub-networks with different widths in each stage of the supernet.

In previous dynamic networks~\cite{veit2018AIG,wang2018skipnet,li2020DynamicRouting,Yang2019CondConvCP,Bolukbasi2017AdaptiveNN, Huang2018MultiScaleDN,Li2019ImprovedTF,hua2019channel,gao2018dynamic,Chen2019YouLT}, the dynamic routing agent and the main network are jointly trained, analogous to jointly optimized NAS methods~\cite{Liu2018DARTSDA, Cai2018ProxylessNASDN, Wu2018FBNetHE}. Inspired by one-shot NAS methods \cite{brock2017smash, akimoto2019adaptive, bender2018understanding, Guo2019SinglePO, li2019blockwisely}, we propose a disentangled two-stage training scheme to ensure the generality of every path in our DS-Net. In \textbf{Stage I}, we disable the slimming gate and train the supernet with the IEB technique, then in \textbf{Stage II}, we fix the weights of the supernet and train the slimming gate with the SGS technique.
\vspace{-5pt}
\subsection{Dynamic Supernet}
\vspace{-5pt}
In this section, we first introduce the hardware efficient channel slicing scheme and our designed supernet, then present the IEB technique and details of training Stage I.

\noindent\textbf{Supernet and Dynamic Channel Slicing.}\label{sect:dyn_ch_slicing}
In some of dynamic networks, such as dynamic pruning~\cite{hua2019channel, gao2018dynamic} and conditional convolution~\cite{Yang2019CondConvCP,li2021revisiting}, the convolution filters $\mathcal{W}$ are conditionally parameterized by a function $\mathcal{A}(\theta, \mathcal{X})$ to the input $\mathcal{X}$. Generally, the dynamic convolution has a form of:
\vspace{-5pt}\begin{equation}
\small
    \mathcal{Y} =  \mathcal{W}_{\mathcal{A}(\theta, \mathcal{X})} * \mathcal{X},
\vspace{-5pt}\end{equation}
where $\mathcal{W}_{\mathcal{A}(\theta, \mathcal{X})}$ represents the selected or generated input-dependent convolution filters. Here $*$ is used to denote a matrix multiplication. Previous dynamic pruning methods \cite{hua2019channel, gao2018dynamic} reduce theoretical computation cost by varying the channel sparsity pattern according to the input. However, they fail to achieve real-world acceleration because their hardware-incompatible channel sparsity results in repeatedly indexing and copying selected filters to a new contiguous memory for multiplication. To achieve practical acceleration, filters should remain contiguous and relatively static during dynamic weight selection.
Base on this analysis, we design a architecture routing agent $\mathcal{A}(\theta)$ with the inductive bias of always outputting a \textit{dense} architecture, \textit{e.g.} a \textit{slice-able} architecture. Specifically, we consider a convolutional layer with at most $N$ output filters and $M$ input channels. Omitting the spatial dimension, its filters can be denoted as $\mathbf{W} \in \mathbb{R}^{N \times M}$.
The output of the architecture routing agent $\mathcal{A}(\theta)$ for this convolution would be a slimming ratio $\rho \in (0,1]$ indicating that the first \textbf{piece-wise} $\rho \times N$ of the output filters are selected. Then, a dynamic slice-able convolution is defined as follows:
\vspace{-5pt}\begin{equation}\label{equ:dyn_slim}
\small
    \mathcal{Y} = \mathbf{W}[~:\rho \times N] * \mathcal{X},
\vspace{-7pt}\end{equation}where $[~:~]$ is a slice operation denoted in a python-like style. \textbf{Remarkably}, the slice operation $[~:~]$ and the dense matrix multiplication $*$ are much more efficient than an indexing operation or a sparse matrix multiplication in real-world implementation, which guarantees a practical acceleration of using our slice-able convolution.

After aggregating the slice-able convolutions sequentially, a supernet executable at different widths is formed. Paths with different widths can be seen as sub-networks. By disabling the routing agent, the supernet is analogous to a slimmable network \cite{Yu2019SlimmableNN, Yu2019UniversallySN}, and can be trained similarly.


\begin{figure}
\vspace{-10pt}
  \centering
   \includegraphics[width=1.0\linewidth]{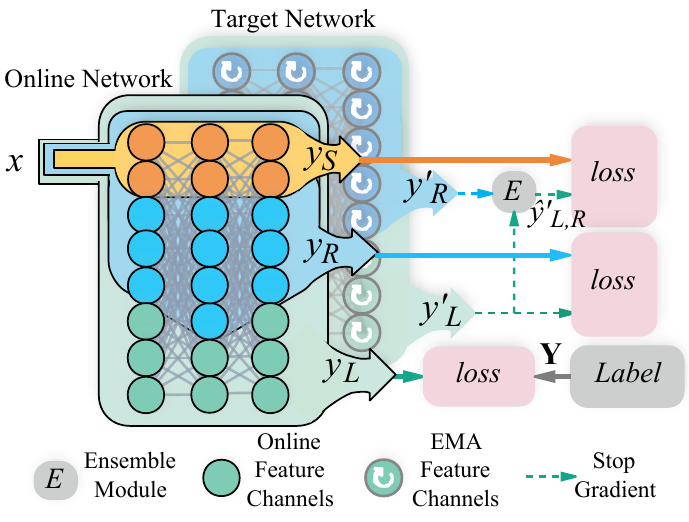}
   \vspace{-15pt}
  \caption{Training process of slimmable supernet with In-place Ensemble Bootstrapping.}\label{fig:ensemble}
  \vspace{-15pt}
\end{figure}
\noindent\textbf{In-place Ensemble Bootstrapping.}
The \emph{sandwich rule} and \emph{in-place distillation} techniques \cite{Yu2019UniversallySN} proposed for Universally Slimmable Networks enhanced their overall performance. 
In \textit{in-place distillation}, the widest sub-network is used as the target network generating soft labels for other sub-networks. However, acute fluctuation appeared in the weight of the widest sub-network can cause convergence hardship, especially in the early stage of training. As observed in BigNAS \cite{Yu2020BigNASSU}, training a more complex model with in-place distillation could be highly unstable. Without residual connection and special weight initialization tricks, the loss exploded at the early stage and can never converge. To overcome the convergence hardship in slimmable networks and improve the overall performance of our supernet, we proposed a training scheme named \textit{In-place Ensemble Bootstrapping (IEB)}.

In recent years, a growing number of self-supervised methods with bootstrapping \cite{BYOL, guo2020bootstrap, deepcluster} and semi-supervised methods based on consistency regularization \cite{laine2016temporal,tarvainen2017mean} use their historical representations to produce targets for the online network.
Inspired by this, we propose to bootstrap on previous representations in our \textit{supervised} in-place distillation training. We use the exponential moving average (EMA) of the model as the target network that generates soft labels. Let $\theta$ and $\theta'$ denote the parameters of the online network and the target network, respectively. We have:\vspace{-5pt}\begin{equation}
\begin{aligned}
    \theta_t' = \alpha\theta_{t-1}' + (1-\alpha)\theta_t,
\end{aligned}
\vspace{-5pt}\end{equation}
where $\alpha$ is a momentum factor controlling the ratio of the historical parameter and $t$ is a training timestamp which is usually measured by a training iteration. During training, the EMA of the model are more stable and more precise than the online network, thus can provide high quality target for the slimmer sub-networks.

As pointed out in \cite{Meal, Mealv2}, an ensemble of teacher networks can generate more diverse, more accurate and more general soft labels for distillation training of the student network. In our supernet, there are tons of sub-models with different architectures, which can generate different soft labels. Motivated by this, we use different sub-networks as a teacher ensemble when performing in-place distillation. The overall train process is shown in Fig. \ref{fig:ensemble}. Following the sandwich rule \cite{Yu2019UniversallySN}, the widest (denoted with $L$), the slimmest (denoted with $S$) and $n$ random width sub-networks (denoted with $R$) are sampled in each training step. Sub-network at the largest width is trained to predict the ground truth label $\mathbf{Y}$; $n$ sub-networks with random width are trained to predict the soft label generated by the widest sub-network of the target network, $\mathcal{Y}_{L}'(\theta')$; the slimmest sub-network is trained to predict the probability ensemble of all the aforementioned sub-networks in the target network:
\vspace{-5pt}\begin{equation}
\footnotesize
\begin{aligned}
    \widehat{\mathcal{Y'}}_{L,R}(\theta') = \frac{1}{n+1}\left(\mathcal{Y}_{L}'(\theta')+ \sum_{i=1}^n\mathcal{Y}_{R}'(\theta')\right).
\end{aligned}
\vspace{-5pt}\end{equation}
To sum up, the IEB losses for the supernet training are:
\begin{equation}
\footnotesize
\left\{
\begin{aligned}
    &\mathcal{L}^{IEB}_{L}(\theta) = \mathcal{L}_{CE}(\mathcal{Y}_{L}(\theta), \mathbf{Y}),\\ 
    &\mathcal{L}^{IEB}_{R}(\theta) = \mathcal{L}_{CE}(\mathcal{Y}_{R}(\theta), \mathcal{Y}_{L}'(\theta')),\\
    &\mathcal{L}^{IEB}_{S}(\theta) = \mathcal{L}_{CE}(\mathcal{Y}_{S}(\theta), \widehat{\mathcal{Y'}}_{L,R}(\theta')),\\
\end{aligned}
\right.
\end{equation}
\subsection{Dynamic Slimming Gate}\label{sec:gate}
\vspace{-5pt}
In this section, we design the channel gate function $\mathcal{A}(\theta, \mathcal{X})$ that generates the factor $\rho$ in Eqn. (\ref{equ:dyn_slim}) and present the \textit{double-headed design} of the dynamic gate. Then, we introduce the details of training stage II with an advanced technique that is \emph{sandwich gate sparsification (SGS)}.



\noindent\textbf{Double-headed Design.}
%
%
There are two possible ways to transform a feature map into a slimming ratio $\rho$ in Eqn. (\ref{equ:dyn_slim}): \textbf{(i) scalar design} directly output a sigmoid activated scalar ranging from 0 to 1 to be the slimming ratio; \textbf{(ii) one-hot design} use an $\mathtt{argmax}$/$\mathtt{softmax}$ activated one-hot vector to choose the respective slimming ratio $\rho$ in a discrete candidate list vector $L_{\rho}$. Both of the implementations are evaluated and compared in Sec. \ref{sec:gate_design}. Here, we thoroughly describe our dynamic slimming gate with the better-performing \textbf{one-hot design}. To reduce the input feature map $\mathcal{X}$ to a one-hot vector, we divide $\mathcal{A}(\theta, \mathcal{X})$ to two functions:
\vspace{-5pt}\begin{equation}
\small
    \mathcal{A}(\theta, \mathcal{X}) = \mathcal{F}(\mathcal{E}(\mathcal{X})),
\vspace{-5pt}\end{equation}
where $\mathcal{E}$ is an encoder that reduces feature maps to a vector and the function $\mathcal{F}$ maps the reduced feature to a one-hot vector used for the subsequent channel slicing. Considering the $n$-th gate in Fig. \ref{fig:network}, given a input feature $\mathcal{X}$ with dimension \begin{small}$\rho_{n-1}C_n \times H_n \times W_n$\end{small}, $\mathcal{E}(\mathcal{X})$ reduces it to a vector \begin{small}$\mathcal{X}_\mathcal{E} \in \mathbb{R}^{\rho_{n-1}C_n}$\end{small} which can be further mapped to a one-hot vector. By computing the dot product of this one-hot vector and $L_{\rho}$, we have the newly predicted slimming ratio:
\vspace{-5pt}\begin{equation}
\small
\rho_n = \mathcal{A}(\theta, \mathcal{X}) \cdot L_{\rho}.
\vspace{-5pt}\end{equation}

Similar to prior works \cite{hu2019squeeze, yang2019gated} on channel attention and gating, we simply utilize average pooling as a light-weight encoder $\mathcal{E}$ to integrate spatial information.
As for feature mapping function $\mathcal{F}$, we adopt two fully connected layers with weights \begin{small}$\mathbf{W}_1 \in \mathbb{R}^{d \times C_n}$\end{small} and \begin{small}$\mathbf{W}_2 \in \mathbb{R}^{g \times d}$\end{small} (where $d$ represents the hidden dimension and $g$ represents the number of candidate slimming ratio) and a ReLU non-linearity layer $\sigma$ in between to predict scores for each slimming ratio choice. An $\mathtt{argmax}$ function is subsequently applied to generate a one-hot vector indicating the predicted choice: \vspace{-5pt}\begin{equation}\label{eqn:att_head}
\small
    \mathcal{F}(\mathcal{X}_\mathcal{E}) = \mathtt{argmax}(\mathbf{W}_2(\sigma(\mathbf{W}_1[:, :\rho_{n-1}C_n](\mathcal{X_\mathcal{E}})))).
\vspace{-5pt}\end{equation}
Note that input $\mathcal{X}$ with dynamic channel number $\rho \times C$ is projected to a vector with fixed length by the dynamically sliced weight \begin{small}$\mathbf{W}_1[:, :\rho_{n-1}C_n]$\end{small}.

Our proposed channel gating function has a similar form with recent channel attention methods \cite{hu2019squeeze, yang2019gated}. The attention mechanism can be integrated into our gate with nearly zero cost, by adding another fully connected layer with weights \begin{small}$\mathbf{W}_3$\end{small} that projects the hidden vector back to the original channel number \begin{small}$\rho_{n-1}C_n$\end{small}. Based on the conception above, we propose a \textit{double-headed dynamic gate} with a soft channel attention head and a hard channel slimming head.
The channel attention head can be defined as follows: 
\vspace{-5pt}\begin{equation}
\small
    \hat{\mathcal{X}} = \mathcal{X} * \delta(\mathbf{W}_3[:\rho_{n-1}C_n, :](\sigma(\mathbf{W}_1[:, :\rho_{n-1}C_n](\mathcal{X})))),
\vspace{-5pt}\end{equation}
where \begin{small}$\delta (x) = 1 + \mathtt{tanh}(x)$\end{small} is the activation function adopted for the attention head.
Unlike the slimming head, the channel attention head is activated in training stage I.

\noindent\textbf{Sandwich Gate Sparsification.}
In training stage II, we propose to use the end-to-end classification cross-entropy loss $\mathcal{L}_{cls}$ and a complexity penalty loss $\mathcal{L}_{cplx}$ to train the gate, aiming to choose the most efficient and effective sub-networks for each instance.
To optimize the non-differentiable slimming head of dynamic gate with $\mathcal{L}_{cls}$, we use $\mathtt{gumbel}$-$\mathtt{softmax}$ \cite{jang2016categorical}, a classical way to optimize neural networks with $\mathtt{argmax}$ by relaxing it to differentiable $\mathtt{softmax}$ in gradient computation. 

However, we empirically found that the gate easily collapses into a static one even if we add Gumbel noise \cite{jang2016categorical} to help the optimization of $\mathtt{gumbel}$-$\mathtt{softmax}$. Apparently, using only $\mathtt{gumbel}$-$\mathtt{softmax}$ technique is not enough for this multi-objective dynamic gate training. 
To further overcome the convergence hardship and increase the dynamic diversity of the gate, a technique named \emph{Sandwich Gate Sparsification (SGS)} is further proposed. We use the slimmest sub-network and the whole network to identify easy and hard samples online and further generate the ground truth slimming factors for the slimming heads of all the dynamic gates.

As analysed in \cite{Yu2019UniversallySN}, wider sub-networks should always be more accurate because the accuracy of slimmer ones can always be achieved by learning new connections to zeros. Thus, given a well-trained supernet, input samples can be roughly classified into three difficulty levels: \textbf{a) Easy samples $\mathcal{X}_{easy}$} that can be correctly classified by the slimmest sub-network; \textbf{b) Hard samples $\mathcal{X}_{hard}$} that can not be correctly classified by the widest sub-network; \textbf{c) Dependent samples $\mathcal{X}_{dep}$}: Other samples in between. In order to minimize the computation cost, easy samples should always be routed to the slimmest sub-network (\textit{i.e.} gate target \begin{small}$\mathcal{T}(\mathcal{X}_{easy}) = [1,0, \dots, 0]$\end{small}). For dependent samples and hard samples, we always encourage them to pass through the widest sub-network, even if the hard samples can not be correctly classified (\textit{i.e.} \begin{small}$\mathcal{T}(\mathcal{X}_{hard}) = \mathcal{T}(\mathcal{X}_{dep}) = [0, \dots, 0, 1]$\end{small}). Another gate target strategy is also discussed in Sec. \ref{sec:SGS_strategy}.

Based on the generated gate target, we define the SGS loss that facilitates the gate training:
\begin{equation}
\footnotesize
\begin{aligned}
    \mathcal{L}_{SGS} = &~\mathbb{T}_{slim}(\mathcal{X})*\mathcal{L}_{CE}(\mathcal{X}, \mathcal{T}(\mathcal{X}_{easy}))\\ 
    &+(\neg\mathbb{T}_{slim}(\mathcal{X}))*\mathcal{L}_{CE}(\mathcal{X}, \mathcal{T}(\mathcal{X}_{hard}))
\end{aligned}
\end{equation}
where \begin{small}$\mathbb{T}_{slim}(\mathcal{X}) \in \{0,1\}$\end{small} represents whether $\mathcal{X}$ is truely predicted by the slimmest sub-network
and \begin{small}$\mathcal{L}_{CE}(\mathcal{X}, \mathcal{T}) = -\sum \mathcal{T} * \log(\mathcal{X})$\end{small} is the Cross-Entropy loss over $\mathtt{softmax}$ activated gate scores and the generated gate target.
\vspace{-5pt}
\section{Experiments}
\vspace{-5pt}
\noindent\textbf{Dataset.} We evaluate our method on two classification datasets (\textit{i.e.}, ImageNet \cite{deng2009imagenet} and CIFAR-10 \cite{Krizhevsky09cifar}) and a standard object detection dataset (\textit{i.e.}, PASCAL VOC \cite{everingham2010pascal}). The ImageNet dataset is a large-scale dataset containing 1.2 M $\mathtt{train}$ set images and 50 K $\mathtt{val}$ set images in 1000 classes. We use all the training data in both of the two training stages. Our results are obtained on the $\mathtt{val}$ set with image size of $224\times224$. We also test the transferability of our DS-Net on CIFAR-10, which comprises 10 classes with 50,000 training and 10,000 test images. Note that few previous works on dynamic networks and network pruning reported results on object detection. We take PASCAL VOC, one of the standard datasets for evaluating object detection performance, as an example to further test the generality of our dynamic networks on object detection. All the detection models are trained with the combined dataset from 2007 $\mathtt{trainval}$ and 2012 $\mathtt{trainval}$ and tested on VOC 2007 $\mathtt{test}$ set.

\noindent\textbf{Architecture details.} Following previous works on static and dynamic network pruning, we use two representative networks, \textit{i.e.}, the heavy residual network ResNet 50 \cite{he2016deep} and the lightweight non-residual network MobileNetV1 \cite{howard2017mobilenets}, to evaluate our method.

In Dynamic Slimmable ResNet 50 \textbf{(DS-ResNet)}, we insert our double-headed gate in the begining of each residual blocks. The slimming head is only used in the first block of each stage. Each one of those blocks contains a skip connection with a projection layer, \textit{i.e.} $1\times1$ convolution. The filter number of this projection convolution is also controlled by the gate to avoid channel inconsistency when adding skip features with residual output. In other residual blocks, the slimming heads of the gates are disabled and all the layers in those blocks inherit the widths of the first blocks of each stage. To sum up, there are 4 gates (one for each stage) with both heads enabled. Every gates have 4 equispaced candidate slimming ratios, \textit{i.e.} $\rho \in \{0.25, 0.5, 0.75, 1\}$. The total routing space contains $4^4 = 256$ possible paths with different computation complexities. All batch normalization (BN) layers in DS-ResNet are replaced with group normalization to avoid test-time representation shift caused by inaccurate BN statistics in weight-sharing networks \cite{Yu2019SlimmableNN,Yu2019UniversallySN}.

Unlike DS-ResNet, we only use one single slimming gate after the fifth depthwise separable convolution block of Dynamic Slimmable MobileNetV1 \textbf{(DS-MBNet)}. Specifically, a fixed slimming ratio $\rho = 0.5$ is used in the first 5 blocks while the width of the rest 8 blocks are controlled by the gate with the candidate slimming ratios $\rho \in[0.35:0.05:1.25]$. This architecture with only 18 paths in its routing space is similar to an uniform slimmable network \cite{Yu2019SlimmableNN,Yu2019UniversallySN}, guaranteeing itself the practicality to use batch normalization. Following \cite{Yu2019UniversallySN}, we perform BN recalibration for all the 18 paths in DS-MBNet after the supernet training stage.

\noindent\textbf{Training details.}
We train our supernet with 512 total batch size on ImageNet, using SGD optimizer with 0.2 initial learning rate for DS-ResNet and 0.08 initial learning rate for DS-MBNet, respectively. We use cosine learning rate scheduler to reduce the learning rate to its 1\% in 150 epochs. Other settings are following previous works on slimmable networks \cite{Yu2019SlimmableNN,Yu2019UniversallySN,yu2019autoslim}.
For gate training, we use SGD optimizer with 0.05 initial learning rate for a total batch size of 512. The learning rate decays to 0.9$\times$ of its value in every epoch. It takes 10 epochs for the gate to converge.
For transfer learning experiments on CIFAR-10, we follow similar settings with \cite{Kornblith2018DoBI} and \cite{Huang2019GPipeET}. We transfer our supernet for 70 epochs including 15 warm-up epochs and use cosine learning rate scheduler with an initial learning rate of 0.7 for a total batch size of 1024. 
For object detection task , we train all the networks following \cite{liu2016ssd} and \cite{li2017fssd} with a total batch size of 128 for 300 epochs. The learning rate is set to 0.004 at the first, then divided by 10 at epoch 200 and 250.

\begin{table}
  \caption{Comparison of state-of-the-art efficient inference methods on ImageNet. \textcolor{brown}{Brown} denotes network pruning methods, \textcolor{cyan}{Blue} denotes dynamic inference methods, \textcolor{orange}{Orange} denotes architecture search methods and \textcolor{purple}{Purple} denotes our method.
  }
  \label{tab:imagenet}
  \centering
  \scriptsize
  \begin{tabular}{>{\centering\arraybackslash}m{0.7cm}|p{2.6cm}|>{\centering\arraybackslash}m{1.0cm}|>{\centering\arraybackslash}m{1.2cm}}
    \toprule
    \multicolumn{2}{c}{Method}      & MAdds & Top-1 Acc.  \\
    \toprule
    \multirow{6}{*}{\makecell{3B\\MAdds}}
    & \textcolor{brown}{SFP} \cite{SoftFilterPruning}                                    & 2.9B & 75.1 \\
    & \textcolor{brown}{ThiNet-70} \cite{luo2017thinet, liu2018RethinkingPruning}        & 2.9B & 75.8 \\
    & \textcolor{brown}{MetaPruning 0.85} \cite{Liu2019MetaPruningML}                    & 3.0B & 76.2 \\
    & \textcolor{cyan}{ConvNet-AIG-50} \cite{veit2018AIG}                               & 3.1B & 76.2 \\
    & \textcolor{orange}{AutoSlim} \cite{yu2019autoslim}                                & 3.0B & 76.0\\
    & \textcolor{purple}{DS-ResNet-L (Ours)}                                              & 3.1B & \textbf{76.6}\\
    \midrule
    \multirow{8}{*}{\makecell{2B\\MAdds}}
    & ResNet-50 0.75$\times$ \cite{he2016deep}                                          & 2.3B & 74.9 \\
    & S-ResNet-50 \cite{Yu2019SlimmableNN}                                              & 2.3B & 74.9 \\
    & \textcolor{brown}{ThiNet-50} \cite{luo2017thinet, liu2018RethinkingPruning}        & 2.1B & 74.7 \\
    & \textcolor{brown}{CP} \cite{He2017ChannelPF}                                       & 2.0B & 73.3 \\
    & \textcolor{brown}{MetaPruning 0.75} \cite{Liu2019MetaPruningML}                    & 2.0B & 75.4 \\
    & \textcolor{cyan}{MSDNet} \cite{Huang2018MultiScaleDN}                             & 2.0B & 75.5 \\
    & \textcolor{orange}{AutoSlim} \cite{yu2019autoslim}                                & 2.0B & 75.6\\
    &  \textcolor{purple}{DS-ResNet-M (Ours)}                                             & 2.2B & \textbf{76.1}\\
    \midrule
    \multirow{5}{*}{\makecell{1B\\MAdds}}
    & ResNet-50 0.5$\times$ \cite{he2016deep}                                           & 1.1B & 72.1 \\
    & \textcolor{brown}{ThiNet-30}  \cite{luo2017thinet, liu2018RethinkingPruning}       & 1.2B & 72.1 \\
    & \textcolor{brown}{MetaPruning 0.5} \cite{Liu2019MetaPruningML}                     & 1.0B & 73.4 \\
    & \textcolor{cyan}{GFNet} \cite{wang2020glance}                                     & 1.2B & 73.8\\
    &  \textcolor{purple}{DS-ResNet-S (Ours)}                                             & 1.2B & \textbf{74.6}\\
    \end{tabular}
    \begin{tabular}{>{\centering\arraybackslash}m{0.7cm}|p{2.6cm}|>{\centering\arraybackslash}m{0.6cm}|>{\centering\arraybackslash}m{0.7cm}|>{\centering\arraybackslash}m{1.2cm}}
    \toprule
    \multicolumn{2}{c}{Method}      & MAdds & Latency & Top-1 Acc.  \\
    \toprule
    \multirow{4}{*}{\makecell{500M\\MAdds}}
    & MBNetV1 1.0$\times$ \cite{howard2017mobilenets}                                   & 569M & 63ms & 70.9 \\
    & US-MBNetV1 1.0$\times$ \cite{Yu2019UniversallySN}                                 & 569M & - & 71.8 \\
    & \textcolor{orange}{AutoSlim} \cite{yu2019autoslim}                                & 572M & - & 73.0 \\
    & \textcolor{purple}{DS-MBNet-L (Ours)}                                             & 565M & 69ms & \textbf{74.5}\\
    \midrule
    \multirow{7}{*}{\makecell{300M\\MAdds}}
    & MBNetV1 0.75$\times$ \cite{howard2017mobilenets}                                  & 317M & 48ms & 68.4 \\
    & US-MBNetV1 0.75$\times$ \cite{Yu2019UniversallySN}                                & 317M & - & 69.5 \\
    & \textcolor{brown}{NetAdapt}  \cite{yang2018netadapt}                               & 284M & - & 69.1 \\
    & \textcolor{brown}{Meta-Pruning} \cite{Liu2019MetaPruningML}                        & 281M & - & 70.6 \\
    & \textcolor{brown}{EagleEye} \cite{li2020eagleeye}                                  & 284M & - & 70.9 \\
    & \textcolor{cyan}{CG-Net-A}  \cite{hua2019channel}                                 & 303M & - & 70.3 \\
    & \textcolor{orange}{AutoSlim} \cite{yu2019autoslim}                                & 325M & - & 71.5 \\
    & \textcolor{purple}{DS-MBNet-M (Ours)}                                             & 319M & 54ms & \textbf{72.8}\\
    \midrule
    \multirow{4}{*}{\makecell{150M\\MAdds}}
    & MBNetV1 0.5$\times$ \cite{howard2017mobilenets}                                   & 150M & 33ms & 63.3 \\
    & US-MBNetV1 0.5$\times$ \cite{Yu2019UniversallySN}                                 & 150M & - & 64.2 \\
    & \textcolor{orange}{AutoSlim} \cite{yu2019autoslim}                                & 150M & - & 67.9 \\
    & \textcolor{purple}{DS-MBNet-S (Ours)}                                             & 153M & 39ms & \textbf{70.1}\\
    
    \bottomrule
  \end{tabular}
  \vspace{-20pt}
\end{table}
\vspace{-5pt}
\subsection{Main Results on ImageNet}
\begin{figure}[t]
\vspace{-10pt}
    \centering
    \includegraphics[width=1.\linewidth]{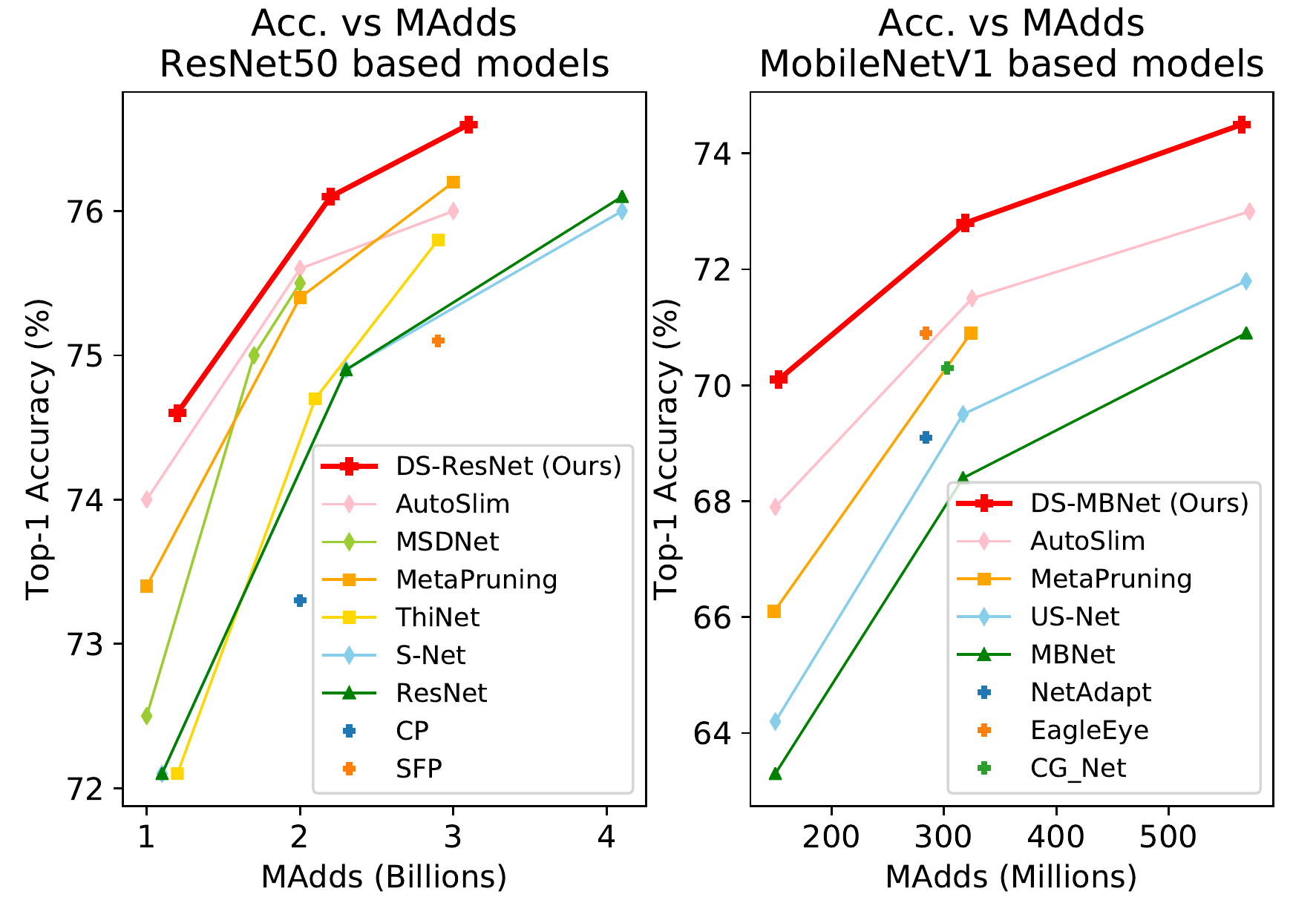}\vspace{-2pt}
    \caption{Accuracy \textit{vs.} complexity on ImageNet.}
    \label{fig:imagenet}
\vspace{-5pt}
\end{figure}

\vspace{-5pt}
We first validate the effectiveness of our method on ImageNet. As shown in Tab. \ref{tab:imagenet} and Fig. \ref{fig:imagenet}, DS-Net with different computation complexity consistently outperforms recent static pruning methods, dynamic inference methods and NAS methods.
\textbf{First}, our DS-ResNet and DS-MBNet models achieve 2-4$\times$ computation reduction over ResNet-50 (76.1\% \cite{he2016deep}) and MobileNetV1 (70.9\% \cite{howard2017mobilenets}) with minimal accuracy drops (0\% to -1.5\% for ResNet and +0.9\% to -0.8\% for MobileNet). We also tested the real world latency on efficient networks. Compare to the ideal acceleration tested on channel scaled MobileNetV1, which is 1.31$\times$ and 1.91$\times$, our DS-MBNet achieves comparable 1.17$\times$ and 1.62$\times$ acceleration with \textit{much} higher performance.
In paticular, our DS-MBNet surpasses the original and the channel scaled MobileNetV1 \cite{howard2017mobilenets} by \textbf{3.6\%}, \textbf{4.4\%} and \textbf{6.8\%} with similar MAdds and minor increase in latency.
\textbf{Second}, our method outperforms classic and state-of-the-art static pruning methods in a large range.
Remarkably, DS-MBNet outperforms the SOTA pruning methods EagleEye \cite{li2020eagleeye} and Meta-Pruning \cite{Liu2019MetaPruningML} by \textbf{1.9\%} and \textbf{2.2\%}. 
\textbf{Third}, our DS-Net maintain superiority comparing with powerful dynamic inference methods with varying depth, width or input resolution. For example, our DS-MBNet-M surpasses dynamic pruning method CG-Net \cite{hua2019channel} by \textbf{2.5\%}.
\textbf{Fourth}, our DS-Net also consistently outperforms its static counterparts. Our DS-MBNet-S surpasses AutoSlim \cite{yu2019autoslim} and US-Net \cite{Yu2019UniversallySN} by \textbf{2.2\%} and \textbf{5.9\%}.

\begin{table}[t]
  \caption{Comparison of transfer learning performance on CIFAR-10. GT stands for gate transfer. 
  }
  \label{tab:cifar}
  \centering
  \footnotesize
  \begin{tabular}{lcc}
    \toprule
    Model                       & MAdds     & Top-1 Acc. \\
    \midrule
    ResNet-50 \cite{he2016deep,Kornblith2018DoBI}& 4.1B      & 96.8  \\
    ResNet-101 \cite{he2016deep,Kornblith2018DoBI} & 7.8B         & 97.6\\
    DS-ResNet w/o GT               & 1.7B      & 97.4  \\
    DS-ResNet w/ GT                & 1.6B      & 97.8  \\

    \bottomrule
  \end{tabular}
  \vspace{-5pt}
\end{table}
\begin{table}[t]
  \caption{Performance comparision of DS-MBNet and MobileNet with FSSD on VOC object detection task.}
  \label{tab:voc}
  \centering
  \footnotesize
  \begin{tabular}{lcc}
    \toprule
    Model                   & MAdds   & mAP \\
    \midrule
    MBNetV1 + FSSD \cite{howard2017mobilenets,li2017fssd}         & 4.3B             & 71.9      \\
    DS-MBNet-S + FSSD      & 2.3B              & 70.7   \\
    DS-MBNet-M + FSSD      & 2.7B              & 72.8\\
    DS-MBNet-L + FSSD      & 3.2B              & \textbf{73.7}\\
    \bottomrule
  \end{tabular}
  \vspace{-10pt}
\end{table}
\vspace{-5pt}
\subsection{Transferability}
\vspace{-5pt}
To evaluate the transferability of DS-Net and its dynamic gate, we perform transfer learning in two settings: \textbf{(i) DS-Net w/o gate transfer}: we transfer the supernet without slimming gate to CIFAR-10 and retrain the dynamic gate. \textbf{(ii) DS-Net w/ gate transfer}: we first transfer the supernet then load the ImageNet trained gate and perform transfer leaning for the gate. The results along with the transfer learning results of the original ResNets are shown in Tab. \ref{tab:cifar}. Gate transfer boosts the performance of DS-ResNet by 0.4\% on CIFAR-10, demonstrating the transferability of dynamic gate. \textbf{Remarkably}, both of our transferred DS-ResNet outperforms the original ResNet-50 in a large range (0.6\% and 1.0\%) with about 2.5 $\times$ computation reduction. Among them, DS-ResNet with gate transfer even outperforms the larger ResNet-101 with 4.9$\times$ fewer computation complexity, proving the superiority of DS-Net in transfer learning.

\vspace{-5pt}
\subsection{Object Detection}
\vspace{-5pt}
In this section, we evaluate and compare the performance of original MobileNet and DS-MBNet used as feature extractor in object detection with Feature Fusion Single Shot Multibox Detector(FSSD) \cite{li2017fssd}. We use the features from the 5-th, 11-th and 13-th depthwise convolution blocks (with the output stride of 8, 16, 32) of MobileNet for the detector. When using DS-MBNet as the backbone, all the features from dynamic source layers are projected to a fixed channel dimention by the feature transform module in FSSD \cite{li2017fssd}.

Results on VOC 2007 $\mathtt{test}$ set are given in Tab. \ref{tab:voc}. Comparing to MobileNetV1, DS-MBNet-M and DS-MBNet-L with FSSD achieves 0.9 and 1.8 mAP improvement with 1.59$\times$ and 1.34$\times$ computation reduction respectively, which demonstrates that our DS-Net remain its superiority after deployed as the backbone network in object detection task.

\begin{figure} \vspace{-15pt}
\centering
{\includegraphics[width=0.9\linewidth]{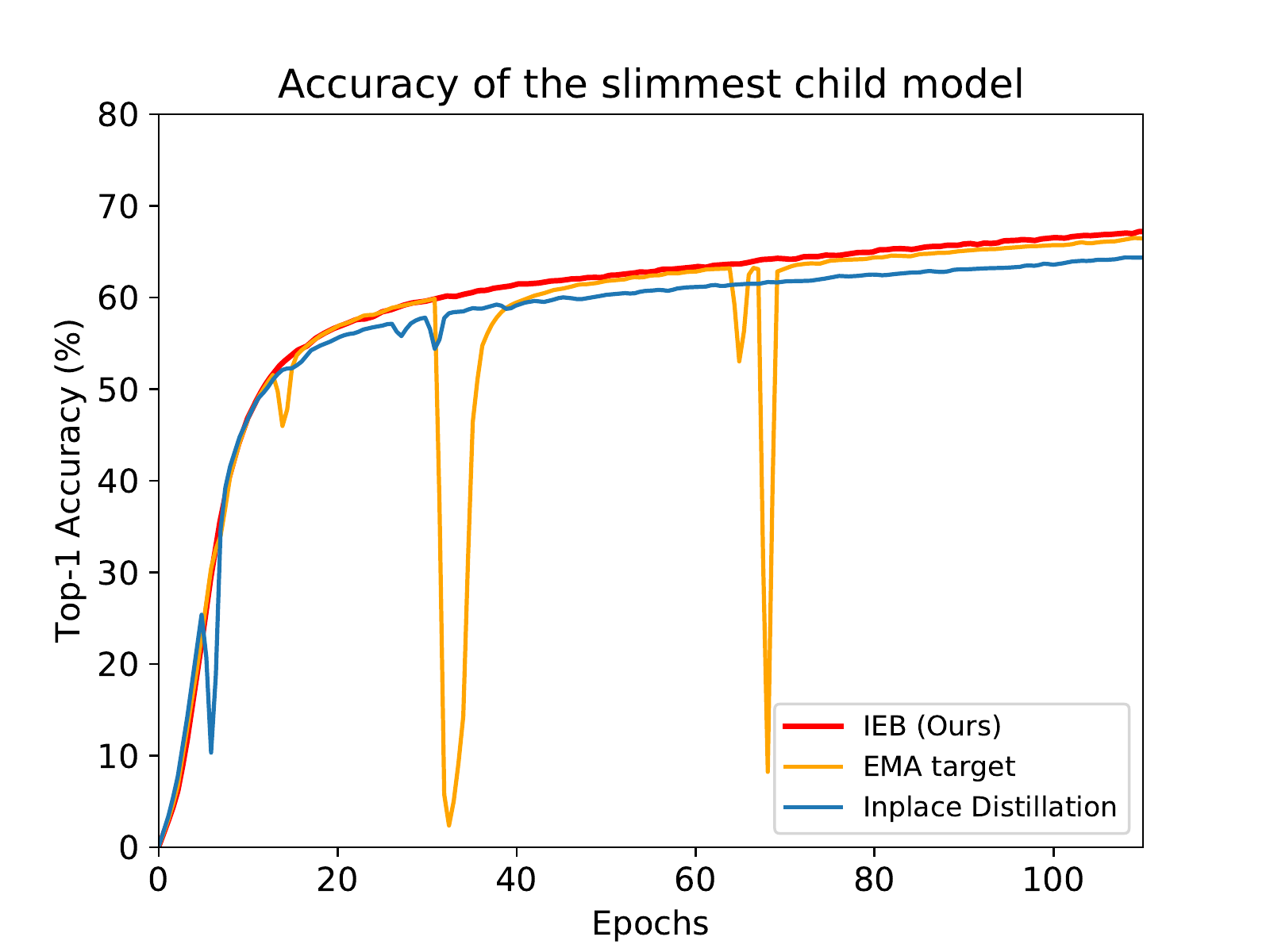}}\vspace{-2pt}
{\caption{Evaluation accuracy of the slimmest sub-network during supernet training with three different training schemes.}\label{fig:ensemble_exp}}
\vspace{-2pt}
\end{figure}
\begin{table}
  \caption{Ablation analysis of In-place Ensemble Bootstrapping.}
  \label{tab:ensemble}
  \centering
  \footnotesize
  \begin{tabular}{cccc}
    \toprule
    EMA                 & Ensemble      & slimmest  & widest \\
    \midrule
                        &               & 66.5      & 74.0\\
     \checkmark         &               & 68.1      & 74.3 \\
     \checkmark         & \checkmark    & 68.3      & 74.6 \\
    \bottomrule
  \end{tabular}
  \vspace{-10pt}
\end{table}

\vspace{-5pt}
\subsection{Ablation study}
\vspace{-5pt}

\noindent\textbf{In-place Ensemble Bootstrapping.}
We statistically analysis the effect of IEB technique with MobileNetV1. We train a Slimmable MobileNetV1 supernet with three settings: original in-place distillation, in-place distillation with EMA target and  our complete IEB technique. As shown in Tab. \ref{tab:ensemble}, the slimmest and widest sub-network trained with EMA target surpassed the baseline by 1.6\% and 0.3\% respectively. With IEB, the supernet improves 1.8\% and 0.6\% on its slimmest and widest sub-networks comparing with in-place distillation. The evaluation accuracy progression curves of the slimmest sub-networks trained with these three settings are illustrated in Fig. \ref{fig:ensemble_exp}. The beginning stage of in-place distillation is unstable. Adopting EMA target improves the performance. However, there are a few sudden drops of accuracy in the middle of the training with EMA target. Though being able to recover in several epochs, the model may still be potentially harmed by those fluctuation. After fully adopting IEB, the model converges to a higher final accuracy without any conspicuous fluctuations in the training process, demonstrating the effectiveness of our IEB technique in stablizing the training and boosting the overall performance of slimmable networks.

\begin{figure}[t]
\vspace{-5pt}
    \centering
    \subfigure{\includegraphics[width=0.45\linewidth]{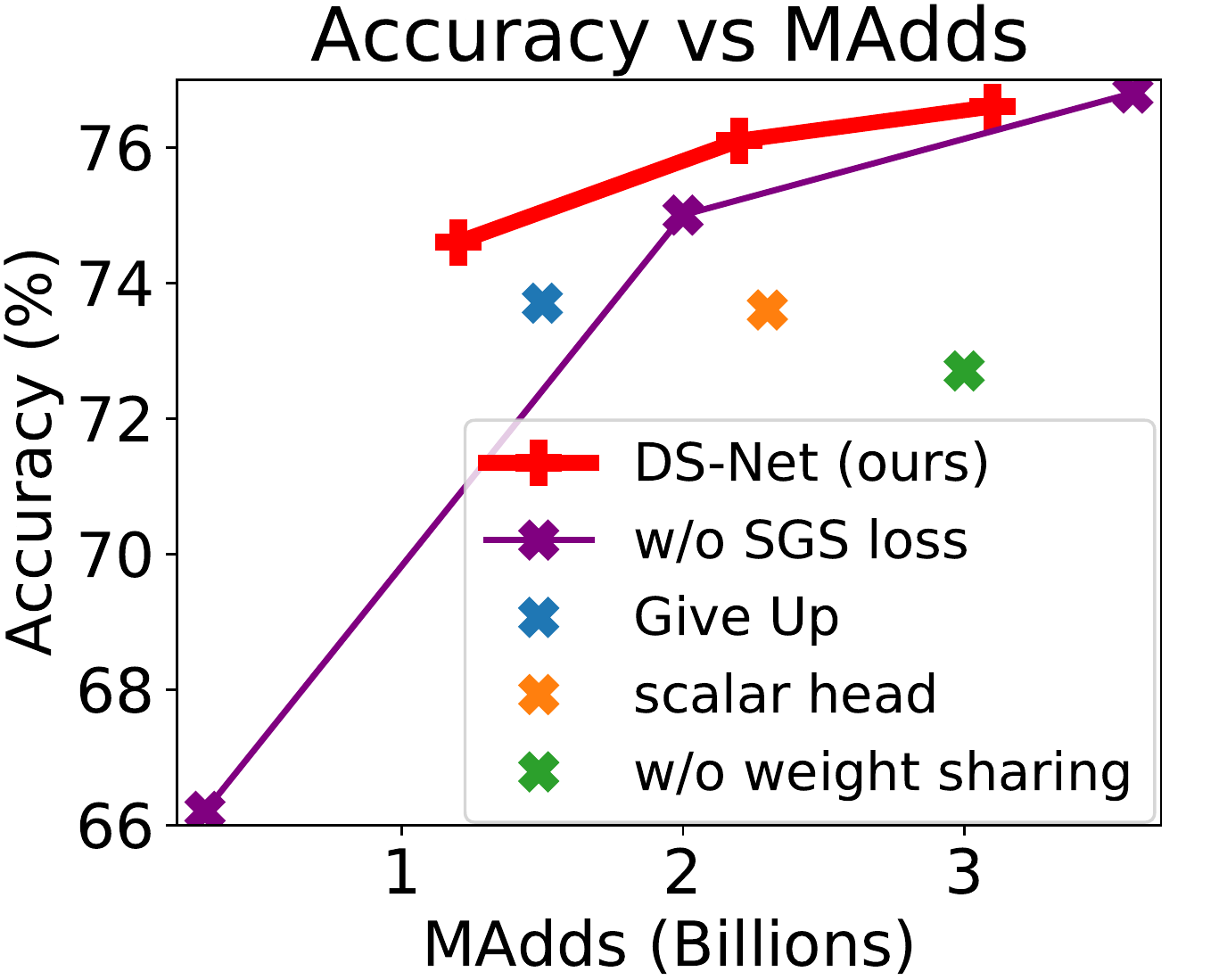}}~~%
    \subfigure{\includegraphics[width=0.55\linewidth]{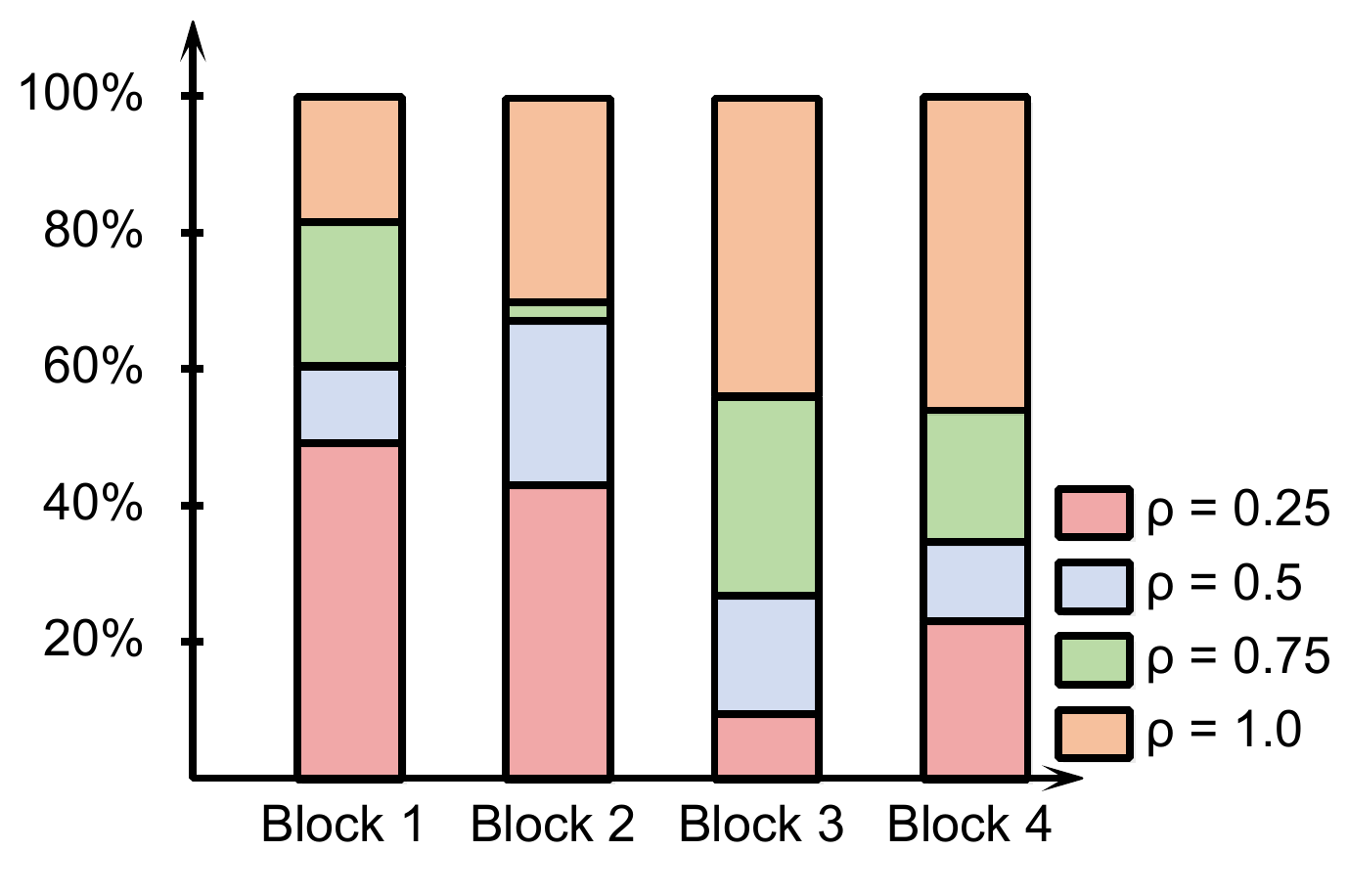}}%

    \caption{\small\textbf{(Left)} Illustration of accuracy \textit{vs.} complexity of models in Tab. \ref{tab:losses} and Tab. \ref{tab:gate_design}. \textbf{(Right)} Gate distribution of DS-ResNet-M. The height of those colored blocks illustrate the partition of input samples that are routed to the sub-networks with respective slimming ratio $\rho$. } 
    \label{fig:gate_distribution}
\vspace{-15pt}
\end{figure}

\noindent\textbf{Effect of losses.}
To examine the impact of the three losses used in our gate training, \textit{i.e.} \emph{target loss} $\mathcal{L}_{cls}$, \emph{complexity loss} $\mathcal{L}_{cplx}$ and \emph{SGS loss} $\mathcal{L}_{SGS}$, we conduct extensive experiments with DS-ResNet on ImageNet, and summarize the results in Tab. \ref{tab:losses} and Fig. \ref{fig:gate_distribution} left. Firstly, as illustrated in Fig. \ref{fig:gate_distribution} left, models trained with SGS (\textcolor{red}{red line}) are more efficient than models trained without it (\textcolor{purple}{purple line}). Secondly, as shown in Tab. \ref{tab:losses}, with target loss, the model pursues better performance while ignoring computation cost; complexity loss pushes the model to be lightweight while ignoring the performance; SGS loss itself can achieve a balanced complexity-accuracy trade-off by encouraging easy and hard samples to use slim and wide sub-networks, respectively.

\noindent
\noindent\textbf{SGS strategy.}\label{sec:SGS_strategy} Though we always want the easy samples to be routed to the slimmest sub-network, there are two possible target definition for hard samples in SGS loss: \textbf{(i) Try Best:} Encourage the hard samples to pass through the widest sub-network, even if they can not be correctly classified (\textit{i.e.} $\mathcal{T}(\mathcal{X}_{hard}) = [0, \dots, 0,1]$).
\textbf{(ii) Give Up:} Push the hard samples to use the slimmest path to save computation cost (\textit{i.e.} $\mathcal{T}(\mathcal{X}_{hard}) = [1,0, \dots, 0]$).
In both of the strategies, dependent samples are encouraged to use the widest sub-network (\textit{i.e.} $\mathcal{T}(\mathcal{X}_{dependent}) = [0, \dots, 0,1]$). The results for both of the strategies are shown in Tab. \ref{tab:losses} and Fig. \ref{fig:gate_distribution} left. As shown in the third and fourth lines in Tab. \ref{tab:losses}, \emph{Give Up} strategy lowers the computation complexity of the DS-ResNet but greatly harms the model performance. The models trained with \emph{Try Best} strategy (\textcolor{red}{red line} in Fig. \ref{fig:gate_distribution} left) outperform the one trained with \emph{Give Up} strategy (\textcolor{cyan}{blue dot} in Fig. \ref{fig:gate_distribution} left) in terms of efficiency. This can be attribute to \emph{Give Up} strategy's optimization difficulty and the lack of samples that targeting on the widest path (dependent samples only 
account for about 10\% of the total training samples).
These results prove our \emph{Try Best} strategy is easier to optimize and can generalize better on validation set or new data.
\begin{table}
  \caption{Ablation analysis of losses on ImageNet. Results in bold that use SGS loss achieve good performance-complexity trade-off.}
  \label{tab:losses}
  \centering
  \footnotesize
  \begin{tabular}{c|c|c|c|c}
    \toprule
    Target      & Complexity        & SGS           & MAdds     & Top-1 Acc.\\
    \midrule
    \checkmark  &                   &               & 3.6B      & 76.8  \\
                & \checkmark        &               & 0.3B      & 66.2  \\
                &                   & \checkmark~~ Give Up       & 1.5B              & 73.7 \\
                &                   & \checkmark~~ Try Best   & \textbf{3.1B}      & \textbf{76.6}  \\
    \checkmark  & \checkmark        &               & 2.0B      & 75.0  \\
                & \checkmark        & \checkmark~~ Try Best   & \textbf{1.2B}      & \textbf{74.6}  \\
    \checkmark  & \checkmark        & \checkmark~~ Try Best   & \textbf{2.2B}      & \textbf{76.1}  \\

    \bottomrule
  \end{tabular}
  \vspace{-5pt}
\end{table}
\begin{table}
  \caption{Ablation analysis of gate design on DS-ResNet.}
  \label{tab:gate_design}
  \centering
  \footnotesize
  \begin{tabular}{>{\centering\arraybackslash}m{1.9cm}>{\centering\arraybackslash}m{1.9cm}>{\centering\arraybackslash}m{0.6cm}>{\centering\arraybackslash}m{1.3cm}}
    \toprule
    weight sharing   & slimming head      & MAdds & Top-1 Acc. \\
    \midrule
    \checkmark      & scalar        & 2.3B  & 73.6 \\
                    & one-hot       & 3.0B  & 72.7 \\
    \checkmark      & one-hot       & 3.1B  & 76.6 \\
    \bottomrule
  \end{tabular}
\vspace{-15pt}
\end{table}

\noindent\textbf{Gate design.}\label{sec:gate_design}
\textbf{First}, to evaluate the effect of our weight-sharing double-headed gate design, we train a DS-ResNet without sharing the the first fully-connected layer for comparison with SGS loss only. As shown in Tab. \ref{tab:gate_design} and Fig. \ref{fig:gate_distribution} left, the performance of DS-ResNet increase substantially (3.9\%) by applying the weight sharing design (\textcolor{green}{green dot} \textit{vs.} \textcolor{red}{red line} in Fig. \ref{fig:gate_distribution} left). This might be attribute to overfitting of the slimming head. As observed in our experiment, sharing the first fully-connected layer with attention head can greatly improve the generality.
\textbf{Second}, we also trained a DS-ResNet with \emph{scalar design} (refer to Sec \ref{sec:gate}) of the slimming head to compare with \emph{one-hot design}. Both of the networks are trained with SGS loss only. The results are present in Tab. \ref{tab:gate_design} and Fig. \ref{fig:gate_distribution} left. The performance of \emph{scalar design} (\textcolor{orange}{orange dot} in Fig. \ref{fig:gate_distribution} left) is much lower than the \emph{one-hot design} (\textcolor{red}{red line} in Fig. \ref{fig:gate_distribution} left), indicating that the scalar gate could not route the input to the correct paths.
\vspace{-5pt}
\subsection{Gate visualization}
\vspace{-5pt}
To demonstrate the dynamic diversity of our DS-Net, we visualize the gate distribution of DS-ResNet over the validation set of ImageNet in Fig. \ref{fig:gate_distribution} right. In block 1 and 2, about half of the inputs are routed to the slimmest sub-network with 0.25 slimming ratio, while in higher level blocks, about half of the inputs are routed to the widest sub-network. For all the gate, the slimming ratio choices are highly input-dependent, demonstrating the high dynamic diversity of our DS-Net.



\vspace{-5pt}
\section{Conclusion}
\vspace{-5pt}
In this paper, we have proposed Dynamic Slimmable Network (DS-Net), a novel dynamic network on efficient inference, achieving good hardware-efficiency by predictively adjusting the filter numbers at test time with respect to different inputs. We propose a two stage training scheme with In-place Ensemble Bootstrapping (IEB) and Sandwich Gate Sparsification (SGS) technique to optimize DS-Net. We demonstrate that DS-Net can achieve 2-4$\times$ computation reduction and 1.62$\times$ real-world acceleration over ResNet-50 and MobileNet with minimal accuracy drops on ImageNet. Proved empirically, DS-Net and can surpass its static counterparts as well as state-of-the-art static and dynamic model compression method on ImageNet by a large margin ($>$2\%) and can generalize well on CIFAR-10 classification task and VOC object detection task.


\vspace{-5pt}
\subsubsection*{Acknowledgments}
\vspace{-5pt}
This work was supported in part by National Key R\&D Program of China under Grant No. 2020AAA0109700, National Natural Science Foundation of China (NSFC) under Grant No.U19A2073, No.61976233 and No.61906109, Guangdong Province Basic and Applied Basic Research (Regional Joint Fund-Key) Grant
No.2019B1515120039, Shenzhen Outstanding Youth Research Project (Project No. RCYX20200714114642083), Shenzhen Basic Research Project (Project No. JCYJ20190807154211365), Leading Innovation Team of the Zhejiang Province (2018R01017) and CSIG Young Fellow Support Fund. Dr Xiaojun Chang is partially supported by the Australian Research Council (ARC) Discovery Early Career Researcher Award (DECRA) (DE190100626).

{\small
\bibliographystyle{ieee_fullname}
\bibliography{egbib}
}

\clearpage
\appendix

\section*{Appendix}

\subsection*{A. Implementation Details}
\noindent\textbf{Losses in Stage II.} Complexity penalty loss $\mathcal{L}_{cplx}$ is used to increase the model efficiency in training stage II. To provide a stable and fair constraint, we use the number of multiply-adds on the fly, $\mathtt{MAdds}(\mathcal{X}, \theta)$, as the metrics of model complexity. Specifically, the complexity penalty is given by:
\begin{equation}
    \small
    \mathcal{L}_{cplx}(\mathcal{X}, \theta) = (\frac{\mathtt{MAdds}(\mathcal{X}, \theta)}{\mathbf{T}})^2,
\end{equation}
where $\mathbf{T}$ is a normalize factor set to the total MAdds of the supernet in our implementation. Note that this loss term always pushes the gate to route towards a faster architecture, towards an architecture with target MAdds, which can effectively prevent routing easy and hard instances to the same architecture.

Overall, the slimming gate can be optimized with a joint loss function:
\begin{equation}
    \small
    \mathcal{L}(\mathcal{X}, \theta) = \lambda_1 \mathcal{L}_{cls} + \lambda_2 \mathcal{L}_{cplx} + \lambda_3 \mathcal{L}_{SGS}.
\end{equation}
The three balancing factors are set to \begin{small}$\lambda_1 = 1$, $\lambda_2 = 0.5$, $\lambda_3 = 1$\end{small} in our experiments. Different target MAdds is reached by adjusting the routing space during gate training. For instance, when training the gate of DS-MBNet-S, we set $\rho \in [0.35:0.05:0.5]$ to prevent routing to heavier sub-networks.

\noindent\textbf{Equispaced channel group.} Following previous works \cite{Yu2019UniversallySN,yu2019autoslim}, we set the the smallest division of channel number to 8. When using $0.05$ as the interval of $\rho$, rounding channels by 8 may result in different intervals, which could lead to training failure when using Group Normalization \cite{Wu2018GroupN}. To prevent such problem, we always adopt a consistent interval (\textit{e.g.} 8, 16, 32) in a single layer, instead of multiplying $\rho$ and rounding the channel. This results in a difference of the slimming ratio between our implemented architecture and our design.

\noindent\textbf{Additional details.}
Weight decay is set to $1^{-4}$ in all of our experiments on ImageNet. To stablize the optimization, weight decay of all the layers in the dynamic gate is removed. The weight $\gamma$ of the last normalization layer of each residual block is initialized to zeros following \cite{Yu2020BigNASSU}. The weight of the fully-connected layer in channel attention head, $\mathbf{W}_3$ in Eqn. \ref{eqn:att_head}, is also zero-initialized to ease the optimization following \cite{yang2019gated}. Additional training techniques include \cite{goyal2017accurate,cubuk2019autoaugment}. We do \textbf{not} use label smoothing \cite{pereyra2017regularizing}, DropPath \cite{larsson2016fractalnet} and RMSProp \cite{tieleman2012lecture}, which are popularly used in previous works \cite{Tan2019EfficientNetRM, Howard2019SearchingFM,yu2019autoslim,Yu2019UniversallySN}.

\subsection*{B. Experiments on EfficientNet}

We also applied our method on EfficientNet \cite{Tan2019EfficientNetRM}, a state-of-the-art network family with high efficiency. Similar to our DS-MBNet, \textbf{D}ynamic \textbf{S}limmable \textbf{Eff}icient\textbf{Net}-B0 (\textbf{DS-EffNet}) has only one slimming gate after its 8-th inverted residual block, controlling the rest 8 blocks. The fixed slimming ratio for the first 8 blocks is 0.5, while a uniform dynamic slimming ratio $\rho \in [0.75:0.05:1.75]$ is used for the last 8 blocks. This supernet with 20 paths in total is trained with a similar config with the supernet of DS-ResNet and DS-MBNet.

We train the supernet with 512 total batch size using 0.2 learning rate that decays with a cosine scheduler in 150 epochs. To enable direct comparision, we opt to reproduce the EfficientNet results using our training setup, with a 150 epoch schedule and no extra enhancement of DropPath \cite{larsson2016fractalnet}, RMSProp \cite{tieleman2012lecture}, \textit{etc.}

The result is shown in Tab. \ref{tab:eff_net}. DS-EffNet outperforms the original EfficientNet-B0 by 0.7\% and 0.8\%, proving its efficacy on recent methods with inverted bottleneck blocks \cite{Sandler2018MobileNetV2IR} and Squeeze-and-Excitation module \cite{hu2019squeeze}.
\begin{table}
  \caption{Comparison of EfficientNet-B0 and DS-EffNet on ImageNet.}
  \label{tab:eff_net}
  \centering
  \footnotesize
  \begin{tabular}{>{\centering\arraybackslash}m{0.9cm}|p{3.0cm}|>{\centering\arraybackslash}m{0.9cm}|>{\centering\arraybackslash}m{1.5cm}}
    \toprule
    \multicolumn{2}{c}{Method}      & MAdds & Top-1 Acc.  \\
    \toprule
    \multirow{2}{*}{\makecell{400M\\MAdds}}
    & EffNet-B0 \cite{Tan2019EfficientNetRM} (repro.)                                      & 399M & 76.0 \\
    & DS-EffNet-L (Ours)                                                                & 400M & \textbf{76.7}\\
    \midrule
    \multirow{2}{*}{\makecell{200M\\MAdds}}
    & EffNet-B0 0.75$\times$ \cite{Tan2019EfficientNetRM}                         & 267M &  74.6\\
    & DS-EffNet-S (Ours)                                                                & 270M & \textbf{75.4}\\
    \bottomrule
    \end{tabular}
\vspace{-5pt}
\end{table}
\vspace{-5pt}

\subsection*{C. Additional Ablations}
\noindent\textbf{Slimming gate.} We analysis the improvement brought by slimming gate by comparing the performance of DS-Net and its supernet. As shown in Tab. \ref{tab:slim_gate}, slimming gate boosts the performance of DS-MBNet-S and DS-ResNet-S by 0.8\% and 1.2\% respectively, comparing to sub-networks with similar sizes in their supernet.

\begin{table}
  \caption{Ablation analysis of slimming gate.}
  \label{tab:slim_gate}
  \centering
  \footnotesize
  \begin{tabular}{lll}
    \toprule
    model      & MAdds & Top-1 Acc. \\
    \midrule
    supernet (DS-MBNet)          & 140M  & 69.3 \\
    DS-MBNet-S         & 153M  & 70.1 \\
    \midrule
    supernet (DS-ResNet)            & 1.1B & 73.4 \\
    DS-ResNet-S         & 1.2B & 74.6 \\
    \bottomrule
  \end{tabular}
\vspace{-5pt}
\end{table}

\noindent\textbf{Distillation temperature.} Temperature $\tau$ in distillation loss was first introduced in \cite{Hinton2015DistillingTK} to control the smoothness of the target. Using a properly larger $\tau$ usually yields better performance of the student. Surprisingly, we find a huge performance degradation in the slimmest sub-network when using larger $\tau$ in in-place distillation. We test $\tau=4$ with DS-MBNet for 40 epochs and compare the it with the performance of default setting ($\tau=1$). As shown in Tab. \ref{tab:temperature}, the performance of the slimmest sub-network decrease by 10.2\% after applying the temperature $\tau=4$.

\begin{table}
  \caption{Ablation analysis of distillation temperature $\tau$ (40 epochs).}
  \label{tab:temperature}
  \centering
  \footnotesize
  \begin{tabular}{lll}
    \toprule
    $\tau$      & slimmest & widest \\
    \midrule
    1           & 59.2  & 65.6 \\
    4           & 49.0  & 67.6 \\
    \bottomrule
  \end{tabular}
\vspace{-15pt}
\end{table}

\end{document}